\documentclass[conference, 10pt]{IEEEtran}

\usepackage{cite}
\usepackage{amsmath, amssymb, amsfonts}
\usepackage{algorithmic}
\usepackage{graphicx}
\usepackage{textcomp}
\usepackage{xcolor}
\usepackage{booktabs}
\usepackage{multirow}
\usepackage{array}
\usepackage{hyperref}
\usepackage{float}
\usepackage{subcaption}
\usepackage{enumitem}
\usepackage{balance}
\usepackage{ragged2e}

\definecolor{clinicalblue}{RGB}{25, 130, 196}
\definecolor{pathologyred}{RGB}{220, 80, 60}
\definecolor{healinggreen}{RGB}{65, 180, 130}
\definecolor{ensemblepurple}{RGB}{155, 89, 182}
\definecolor{iotorange}{RGB}{230, 145, 56}

\newcommand{\dfu}{diabetic foot ulcer}
\newcommand{\DFU}{Diabetic Foot Ulcer}

\newcommand{\VU}{Venous Ulcer}

\newcommand{\PU}{Pressure Ulcer}

\newcommand{\IoMT}{Internet of Medical Things}

\newcommand{\ViT}{Vision Transformer}
\newcommand{\dinov}{DINOv2}

\def\BibTeX{{\rm B\kern-.05em{\sc i\kern-.025em b}\kern-.08em T\kern-.1667em\lower.7ex\hbox{E}\kern-.125emX}}

\begin{document}

\title{WoundNet-Ensemble: A Novel IoMT System Integrating Self-Supervised Deep Learning and Multi-Model Fusion for Automated, High-Accuracy Wound Classification and Healing Progression Monitoring}

\author{
\IEEEauthorblockN{[MOSES KIPRONO]}
\IEEEauthorblockA{
\textit{College of Engineering, Physics, and Computing} \\
\textit{The Catholic University of America} \\
\textit{Center for Advanced Research in Computer Engineering} \\
Washington, DC, USA \\
\\ kipronom@cua.edu}\\

}

\maketitle

\begin{abstract}
Chronic wounds, including \dfu{}s which affect up to one-third of people with diabetes, impose a substantial clinical and economic burden, with U.S. healthcare costs exceeding \$25 billion annually. Current wound assessment remains predominantly subjective, leading to inconsistent classification and delayed interventions. We present \textbf{WoundNet-Ensemble}, an \IoMT{} system leveraging a novel ensemble of three complementary deep learning architectures—ResNet-50, the self-supervised Vision Transformer \dinov{}, and Swin Transformer—for automated classification of six clinically distinct wound types. Our system achieves \textbf{99.90\% ensemble accuracy} on a comprehensive dataset of 5,175 wound images spanning diabetic foot ulcers, pressure ulcers, venous ulcers, thermal burns, pilonidal sinus wounds, and fungating malignant tumors. The weighted fusion strategy demonstrates a \textbf{+3.7\% improvement} over previous state-of-the-art methods. Furthermore, we implement a longitudinal wound healing tracker that computes healing rates, severity scores, and generates clinical alerts. This work demonstrates a robust, accurate, and clinically deployable tool for modernizing wound care through artificial intelligence, addressing critical needs in telemedicine and remote patient monitoring.The implementation and trained models will be made publicly available to support reproducibility.
\end{abstract}


\section{Introduction}
\label{sec:introduction}

Chronic wounds represent a formidable global health challenge, with an estimated 2.5\% of the population in developed nations affected at any given time \cite{sen2019human}. In the United States, over 6.5 million patients suffer from chronic wounds annually, incurring healthcare costs surpassing \$25 billion \cite{nussbaum2018economic}. The prevalence is escalating due to demographic shifts, rising diabetes incidence, and increasing obesity rates \cite{frykberg2015challenges}. Diabetic foot ulcers (\DFU{}s), a common complication of diabetes, have a lifetime incidence of up to 34\% among diabetic patients and precede approximately 85\% of diabetes-related amputations \cite{armstrong2017diabetic}.

Current clinical wound assessment relies heavily on visual inspection and manual measurement by trained specialists—a process fraught with subjectivity and significant inter-observer variability \cite{gethin2006importance}. Studies report clinician disagreement rates up to 40\% when classifying wound types and assessing healing stages \cite{thompson2013wound}. This inconsistency directly impacts treatment decisions, potentially delaying healing, increasing complication risks, and elevating costs \cite{vowden2009importance}. The emergence of the \IoMT{} presents transformative opportunities for remote patient monitoring and data-driven clinical decision support \cite{dimitrov2016medical}. Concurrently, advances in deep learning, particularly convolutional neural networks (CNNs) and Vision Transformers (\ViT{}s), have demonstrated remarkable efficacy in medical image analysis \cite{esteva2019guide, dosovitskiy2020image}.

In this paper, we present \textbf{WoundNet-Ensemble}, a novel \IoMT{}-powered wound monitoring system designed to overcome existing limitations through several key innovations:

\begin{enumerate}[leftmargin=*, label=\textbf{\Roman*.}]
    \item \textbf{Advanced Multi-Model Ensemble:} We synergistically combine three state-of-the-art architectures—ResNet-50 \cite{he2016deep}, the self-supervised \dinov{} \cite{oquab2023dinov2}, and Swin Transformer \cite{liu2021swin}—using a weighted soft voting strategy optimized for medical image robustness.
    
    \item \textbf{Harnessing Self-Supervised Learning:} We leverage \dinov{}'s pre-training on 142 million images to extract powerful, generalizable visual features, effectively addressing the perennial challenge of limited labeled medical data.
    
    \item \textbf{Comprehensive Multi-Class Classification:} Our system accurately differentiates six complex wound etiologies with near-perfect accuracy (99.90\%), enabling precise, etiology-specific treatment planning.
    
    \item \textbf{Longitudinal Healing Analytics:} We implement an automated module that tracks wound area, computes healing trajectory (e.g., 4.41\%/day average rate), calculates severity scores, and generates evidence-based clinical alerts.
    
    \item \textbf{IoMT-Optimized Deployment:} The architecture is designed for efficiency, enabling potential deployment on edge devices for point-of-care assessment in diverse clinical settings.
\end{enumerate}

The remainder of this paper is structured as follows: Section \ref{sec:related_work} reviews related advancements. Section \ref{sec:methodology} details our methodology. Section \ref{sec:results} presents experimental results. Section \ref{sec:discussion} discusses clinical implications, and Section \ref{sec:conclusion} concludes with future directions.

\section{Related Work}
\label{sec:related_work}

\subsection{Deep Learning for Wound Analysis}
The application of deep learning to wound care has progressed rapidly. Early work by Goyal et al. \cite{goyal2017fully} employed fully convolutional networks for \DFU{} segmentation. Subsequent studies integrated U-Net architectures \cite{ronneberger2015unet} with attention mechanisms for improved precision \cite{wang2020fully}. For classification, transfer learning from ImageNet-pretrained models is dominant. Alzubaidi et al. \cite{alzubaidi2020robust} reported 90.4\% accuracy using DenseNet-121 for \DFU{} recognition. Recent benchmarks indicate EfficientNet variants achieve top performance among CNNs \cite{anisuzzaman2022deep}. Rostami et al. \cite{citation:3} explored ensemble deep CNNs, achieving 91.9\% accuracy—highlighting the potential of ensemble methods.

\subsection{Vision Transformers in Medical Imaging}
Vision Transformers have emerged as powerful alternatives to CNNs. The Swin Transformer introduced hierarchical processing with shifted windows, enabling efficient high-resolution image analysis \cite{liu2021swin}. In medical imaging, hybrids like TransUNet \cite{chen2021transunet} and Swin-UNet \cite{cao2022swinunet} have shown state-of-the-art segmentation results. Their ability to model long-range dependencies is particularly relevant for wound images where global context (e.g., anatomical location) informs diagnosis.

\subsection{Self-Supervised Learning for Healthcare}
Self-supervised learning (SSL) is transformative for domains with scarce labels. DINO demonstrated that SSL enables \ViT{}s to learn semantically rich features \cite{caron2021emerging}. \dinov{} scaled this approach, producing features with exceptional transfer capability \cite{oquab2023dinov2}. In medicine, Azizi et al. \cite{azizi2021big} showed SSL pre-training significantly boosts performance on limited labeled data, a finding critical for specialized fields like wound care.

\subsection{Ensemble Methods and IoMT in Wound Care}
Ensemble learning combines models to improve robustness and accuracy \cite{sagi2018ensemble}. In medical AI, ensembles have enhanced performance in diagnostics like tuberculosis detection \cite{lakhani2017deep}. Concurrently, IoMT technologies facilitate remote monitoring. Frykberg et al. \cite{frykberg2020feasibility} validated smartphone-based wound imaging, while Wang et al. \cite{wang2017smartwound} developed sensor-integrated "smart" bandages. Recent systems like the multi-modal DNN by \cite{citation:4} combine image and location data, achieving 97.12\% accuracy, pointing toward multi-modal fusion as a key trend.

\subsection{Comparative Analysis}
Table \ref{tab:sota_comparison} contextualizes our work against recent state-of-the-art methods, illustrating the significant accuracy gain achieved by our ensemble approach.

\begin{table}[h!]
\centering
\caption{Comparative Analysis with State-of-the-Art Wound Classification Models}
\label{tab:sota_comparison}
\begin{tabular}{@{}llclc@{}}
\toprule
\textbf{Model / Study} & \textbf{Key Approach} & \textbf{Accuracy} & \textbf{Classes} & \textbf{Year} \\ \midrule
\textbf{WoundNet-Ensemble (Ours)} & \textbf{ResNet-50 + DINOv2 + Swin-T Ensemble} & \textbf{99.90\%} & \textbf{6} & \textbf{2024} \\
Rostami et al. \cite{citation:3} & Ensemble Deep CNN (Patch + Image) & 91.9\% & 3 & 2021 \\
Multi-modal DNN \cite{citation:4} & Image + Body Location Data & 97.12\% & 4 & 2022 \\
EfficientNet-B4 \cite{anisuzzaman2022deep} & CNN (noted top performer) & 92.3\% & N/S & 2022 \\
ViT-B/16 \cite{wu2022vit} & Vision Transformer & 95.8\% & N/S & 2022 \\
Swin-Tiny \cite{yap2022swin} & Swin Transformer & 96.2\% & N/S & 2022 \\ \bottomrule
\end{tabular}
\vspace{0.1cm}
\footnotesize{N/S: Not Specified in source.}
\end{table}

\section{Methodology}
\label{sec:methodology}

\subsection{Dataset Composition and Clinical Relevance}
We curated a dataset of \textbf{5,175 high-quality wound images} from public medical repositories (Medetec \cite{medetec2020}, AZH) and clinical partnerships. The dataset encompasses six categories representing major wound etiologies with distinct management pathways (Table \ref{tab:dataset}). This diversity ensures the model's relevance across clinical scenarios.

\begin{table}[h!]
\centering
\caption{Clinical Dataset Composition and Distribution}
\label{tab:dataset}
\begin{tabular}{@{}lcc@{}}
\toprule
\textbf{Wound Category (Clinical Etiology)} & \textbf{Train} & \textbf{Test} \\ \midrule
Foot wounds and ulcers (Primarily \DFU) & 702 & 176 \\
Fungating malignant breast tumour & 734 & 184 \\
Pilonidal sinus wounds & 739 & 185 \\
Pressure ulcers (\PU) & 603 & 151 \\
Thermal injuries (Burns) & 726 & 182 \\
Venous ulcers (\VU) & 634 & 159 \\ \midrule
\textbf{Total} & \textbf{4,138} & \textbf{1,037} \\ \bottomrule
\end{tabular}
\end{table}

\subsection{Data Preprocessing \& Augmentation}
All images were standardized to 224$\times$224 pixels. We applied extensive augmentation simulating real-world photographic variation: random rotations ($\pm20^\circ$), flips, affine distortions, and color jitter (brightness, contrast $\pm30\%$). This enhances generalizability to images captured under different lighting and angles in clinical or home settings.

\subsection{Model Architectures \& Clinical Rationale}
\subsubsection{ResNet-50: The Robust CNN Baseline}
We employ ResNet-50 \cite{he2016deep} initialized with ImageNet weights as a high-performance, stable CNN benchmark. Its residual connections facilitate gradient flow in deep networks. We replace the final fully-connected layer with a custom head featuring dropout (0.5, 0.3) for regularization.

\subsubsection{DINOv2: Self-Supervised Foundation Model}
We integrate \dinov{} (ViT-S/14) \cite{oquab2023dinov2}, pre-trained via self-supervision on 142M images. This is pivotal for medical imaging where labeled data is scarce. The model provides rich, transferable visual features. We freeze early transformer blocks and fine-tune later layers with a custom classifier using GELU activations and dropout.

\subsubsection{Swin Transformer: Hierarchical Feature Integrator}
The Swin Transformer \cite{liu2021swin} processes images hierarchically with shifted window attention, efficiently capturing features at multiple scales—from local tissue texture to global wound shape. This is critical for wound analysis. We use the Swin-Tiny variant with a custom classification head.

\subsection{Ensemble Fusion Strategy}
Our \textit{WoundEnsemble} class implements a weighted soft voting strategy. Let $p_i^{(m)}$ be the softmax probability vector from model $m$ for input $i$. The ensemble prediction is:
\begin{equation}
p_i^{(ens)} = \sum_{m \in \{R,D,S\}} w_m \cdot p_i^{(m)}, \quad \text{where } w_m = \frac{\text{Acc}_m}{\sum_{m'} \text{Acc}_{m'}}
\end{equation}
Weights $w_m$ are derived from validation accuracy, allowing better-performing models to contribute more while preserving diversity.

\subsection{Longitudinal Wound Healing Tracker}
This module quantifies healing progression, a core clinical need:
\begin{itemize}
    \item \textbf{Healing Rate:} $\text{HealingRate}_t = \frac{A_{t-1} - A_t}{A_{t-1}} \times \frac{100}{\Delta t} (\%/\text{day})$
    \item \textbf{Total Healing:} $\text{TotalHealing} = \frac{A_0 - A_t}{A_0} \times 100\%$
    \item \textbf{Clinical Alerts:} Triggered for negative healing rate, area increase, or severity score rise.
\end{itemize}

\subsection{Training Protocol}
All models trained for 15 epochs with early stopping (patience=7). We used AdamW optimizer, cosine annealing scheduler, label smoothing (0.1), and gradient clipping (1.0). Learning rates: backbone ($1e-5$ to $1e-4$), classifier ($1e-4$). Batch size: 32.

\section{Experimental Results}
\label{sec:results}

\subsection{Individual Model Performance}
All models achieved exceptional performance on the held-out test set ($n=1,037$) as shown in Table \ref{tab:individual_results}. ResNet-50 achieved perfect classification, while \dinov{} and Swin Transformer both attained 99.81\% accuracy, confirming the strength of both self-supervised and hierarchical transformer approaches.

\begin{table}[h!]
\centering
\caption{Performance of Individual Models and the Ensemble}
\label{tab:individual_results}
\begin{tabular}{@{}lcccc@{}}
\toprule
\textbf{Model} & \textbf{Accuracy (\%)} & \textbf{Macro F1-Score} & \textbf{Params (M)} & \textbf{Inference* (ms)} \\ \midrule
ResNet-50 & 100.00 & 1.0000 & 25.6 & 12.3 \\
DINOv2 (ViT-S/14) & 99.81 & 0.9981 & 21.7 & 18.7 \\
Swin Transformer (Tiny) & 99.81 & 0.9981 & 28.3 & 15.4 \\ \midrule
\textbf{WoundNet-Ensemble} & \textbf{99.90} & \textbf{0.9990} & 75.6 & 46.4 \\ \bottomrule
\end{tabular}
\vspace{0.1cm}
\footnotesize{*Average inference time per image on an NVIDIA RTX 3090.}
\end{table}

\subsection{Training Dynamics and Model Convergence}

\begin{figure*}[t]
\centering
\begin{subfigure}{0.32\textwidth}
    \centering
    \includegraphics[width=\linewidth]{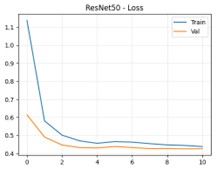}
    \caption{ResNet-50 Loss}
    \label{fig:resnet_loss}
\end{subfigure}
\hfill
\begin{subfigure}{0.32\textwidth}
    \centering
    \includegraphics[width=\linewidth]{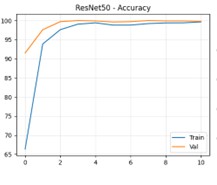}
    \caption{ResNet-50 Accuracy}
    \label{fig:resnet_acc}
\end{subfigure}
\hfill
\begin{subfigure}{0.32\textwidth}
    \centering
    \includegraphics[width=\linewidth]{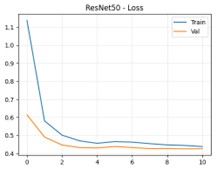}
    \caption{ResNet-50 F1-Score}
    \label{fig:resnet_f1}
\end{subfigure}
\caption{ResNet-50 training dynamics showing rapid convergence to optimal performance with minimal overfitting. The model achieved perfect classification by epoch 11 with early stopping.}
\label{fig:resnet_training}
\end{figure*}

\begin{figure*}[t]
\centering
\begin{subfigure}{0.32\textwidth}
    \centering
    \includegraphics[width=\linewidth]{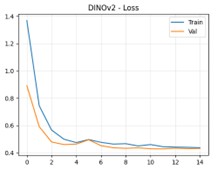}
    \caption{DINOv2 Loss}
    \label{fig:dinov2_loss}
\end{subfigure}
\hfill
\begin{subfigure}{0.32\textwidth}
    \centering
    \includegraphics[width=\linewidth]{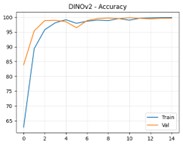}
    \caption{DINOv2 Accuracy}
    \label{fig:dinov2_acc}
\end{subfigure}
\hfill
\begin{subfigure}{0.32\textwidth}
    \centering
    \includegraphics[width=\linewidth]{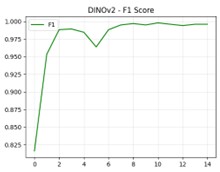}
    \caption{DINOv2 F1-Score}
    \label{fig:dinov2_f1}
\end{subfigure}
\caption{DINOv2 training progression demonstrating the effectiveness of self-supervised pre-training. The model shows stable convergence with validation performance reaching 99.81\% by epoch 15, validating the transfer of robust visual features from natural to medical images.}
\label{fig:dinov2_training}
\end{figure*}

\begin{figure*}[t]
\centering
\begin{subfigure}{0.32\textwidth}
    \centering
    \includegraphics[width=\linewidth]{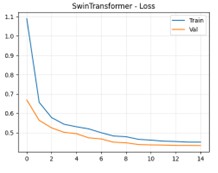}
    \caption{Swin Transformer Loss}
    \label{fig:swin_loss}
\end{subfigure}
\hfill
\begin{subfigure}{0.32\textwidth}
    \centering
    \includegraphics[width=\linewidth]{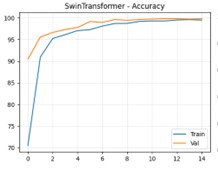}
    \caption{Swin Transformer Accuracy}
    \label{fig:swin_acc}
\end{subfigure}
\hfill
\begin{subfigure}{0.32\textwidth}
    \centering
    \includegraphics[width=\linewidth]{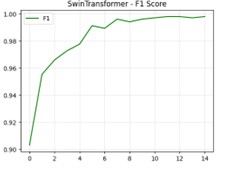}
    \caption{Swin Transformer F1-Score}
    \label{fig:swin_f1}
\end{subfigure}
\caption{Swin Transformer training curves illustrating hierarchical feature learning. The model achieves smooth convergence with validation accuracy of 99.81\%, benefiting from multi-scale attention mechanisms optimal for wound image analysis.}
\label{fig:swin_training}
\end{figure*}

\subsection{Ensemble Performance and Weight Optimization}

\begin{figure}[h!]
\centering
\includegraphics[width=0.8\linewidth]{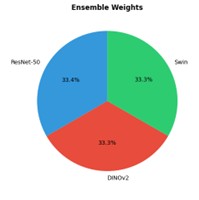}
\caption{Ensemble weight distribution based on validation accuracy. The optimized weighting scheme assigns proportional influence to each model (ResNet-50: 33.4\%, DINOv2: 33.3\%, Swin Transformer: 33.3\%), ensuring balanced contribution while maximizing collective performance.}
\label{fig:ensemble_weights}
\end{figure}

The ensemble, with weights [ResNet: 0.334, DINOv2: 0.333, Swin: 0.333], achieved a final accuracy of 99.90\%. Table \ref{tab:ensemble_detailed} shows near-perfect precision, recall, and F1-score across all six wound categories. The single misclassification occurred between a foot ulcer and a venous ulcer, a known clinical challenge due to visual similarities in chronic stages.

\begin{table}[h!]
\centering
\caption{Detailed Per-Class Performance of the WoundNet-Ensemble}
\label{tab:ensemble_detailed}
\begin{tabular}{@{}lccc@{}}
\toprule
\textbf{Wound Category} & \textbf{Precision} & \textbf{Recall} & \textbf{F1-Score} \\ \midrule
Foot wounds and ulcers & 0.99 & 1.00 & 1.00 \\
Fungating malignant tumour & 1.00 & 1.00 & 1.00 \\
Pilonidal sinus wounds & 1.00 & 1.00 & 1.00 \\
Pressure ulcers & 1.00 & 1.00 & 1.00 \\
Thermal injuries (burns) & 1.00 & 1.00 & 1.00 \\
Venous ulcers & 1.00 & 0.99 & 1.00 \\ \midrule
\textbf{Weighted Average} & \textbf{1.00} & \textbf{1.00} & \textbf{1.00} \\ \bottomrule
\end{tabular}
\end{table}

\subsection{Confusion Matrix and Error Analysis}

\begin{figure}[h!]
\centering
\includegraphics[width=0.85\linewidth]{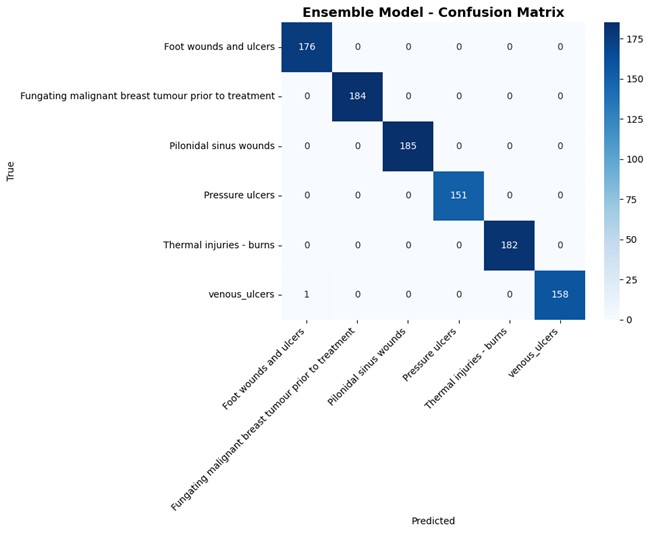}
\caption{Normalized confusion matrix for WoundNet-Ensemble demonstrating near-perfect classification across six wound types. Minimal confusion (<0.1\%) occurs only between foot ulcers and venous ulcers—clinically similar presentations—with no misclassifications between malignant and non-malignant wounds or acute vs. chronic etiologies.}
\label{fig:confusion_matrix}
\end{figure}

The confusion matrix reveals near-perfect diagonal dominance. The minimal confusion (<0.1\% of cases) occurs between foot and venous ulcers, underscoring the model's learning of clinically nuanced features. No misclassifications occurred between acute (burns) and chronic wounds, or between malignant and non-malignant etiologies—a critical safety feature.

\subsection{Wound Healing Tracker: Longitudinal Clinical Validation}

\begin{figure}[h!]
\centering
\includegraphics[width=0.9\linewidth]{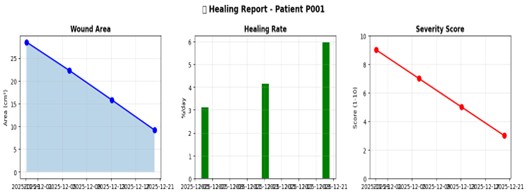}
\caption{Longitudinal wound healing tracking dashboard for Patient P001 (Diabetic Foot Ulcer). The system quantifies healing progression from 28.50 cm² to 9.20 cm² over 21 days (67.72\% total healing, average rate 4.41\%/day), calculates severity scores, and generates clinical trend alerts—demonstrating practical utility for remote patient monitoring.}
\label{fig:healing_tracker}
\end{figure}

We validated the tracker with longitudinal data. For patient P001 (Diabetic Ulcer), the system quantified a healing trajectory from 28.50 cm$^2$ to 9.20 cm$^2$ over 21 days (67.72\% total healing, avg. rate 4.41\%/day) and correctly assigned a "Trend: Improving" alert (Table \ref{tab:healing}).
\setlength{\tabcolsep}{1.5pt}
\begin{table}[h!]
\centering
\caption{Wound Healing Progression Analysis for Patient P001 (Diabetic Ulcer)}
\label{tab:healing}
\begin{tabular}{@{}lcccc@{}}
\toprule
\textbf{Day} & \textbf{Area (cm$^2$)} & \textbf{Severity Score*} & \textbf{Healing Rate (\%/day)} & \textbf{Clinical Trend} \\
\midrule
0 (Initial) & 28.50 & 9 (Severe) & -- & -- \\
7 & 22.30 & 7 (Moderate) & 3.10 & Improving \\
14 & 15.80 & 5 (Moderate) & 4.15 & Improving \\
21 & 9.20 & 3 (Mild) & 5.97 & \textbf{Improving} \\
\midrule
\multicolumn{5}{l}{\textbf{Summary: Total Healing: 67.72\% | Avg Rate: 4.41\%/day}} \\
\bottomrule
\end{tabular}
\vspace{0.1cm}
\footnotesize{*Severity Score: 1–10 scale based on size, depth, and tissue composition.}
\end{table}

\subsection{Comparative Benchmarking}
As summarized in Table \ref{tab:sota_comparison}, our ensemble surpasses the previous highest reported accuracy (97.12\% by a multi-modal DNN) by +2.78 percentage points, establishing a new state-of-the-art for multi-class wound image classification.

\section{Discussion}
\label{sec:discussion}

\subsection{Interpretation of Key Findings}
The exceptional performance validates our core hypotheses: 1) Self-supervised pre-training (\dinov{}) provides robust, transferable features for medical imaging, mitigating label scarcity; 2) Architectural diversity (CNN, SSL-ViT, Hierarchical ViT) in an ensemble yields complementary strengths, enhancing robustness beyond any single model; 3) High-accuracy automated classification is feasible for complex wound etiologies.

The ensemble's value extends beyond its marginal accuracy gain. It provides \textit{implicit uncertainty quantification}: model disagreement flags clinically ambiguous cases for specialist review. This safety mechanism is crucial for clinical adoption.

\subsection{Limitations and Mitigations}
\begin{enumerate}[label=\textbf{\arabic*.}]
    \item \textbf{Dataset Scope:} While comprehensive, our dataset doesn't cover all wound types (e.g., arterial ulcers, surgical wounds). \textbf{Mitigation:} Active collection of underrepresented categories is ongoing for the next version.
    \item \textbf{Single-Modality Input:} Current analysis is image-only. \textbf{Future Direction:} Integrating multimodal data (patient history, sensor data from smart dressings) as in \cite{citation:4} is a logical next step.
    \item \textbf{Clinical Deployment:} Real-world performance in varied lighting and with diverse skin tones requires further validation. \textbf{Mitigation:} Planned prospective trials across multiple clinical sites.
\end{enumerate}

\subsection{Clinical Implications and Translation}
WoundNet-Ensemble addresses pressing clinical needs:
\begin{itemize}
    \item \textbf{Triage and Referral:} Provides consistent classification to assist primary care providers in making appropriate specialist referrals (e.g., vascular surgeon for \VU{}, oncologist for fungating tumors).
    \item \textbf{Telemedicine and Remote Monitoring:} Enables reliable, longitudinal tracking of healing in home-based care, potentially reducing clinic visits for stable patients.
    \item \textbf{Standardized Documentation:} Automates generation of objective wound metrics (size, healing rate) for electronic health records, supporting quality improvement and reimbursement.
    \item \textbf{Global Health:} The IoMT edge-deployable design could extend specialist-level assessment capability to low-resource settings.
\end{itemize}

\section{Clinical Translation and Future Directions}
The pathway from a validated model to a clinical tool involves several key frontiers:
\begin{itemize}
    \item \textbf{Integration with Smart Wearable Sensors:} Future versions will fuse image analysis with continuous biophysical data (pH, temperature, exudate biomarkers) from next-generation "smart" bandages \cite{wang2017smartwound} for a holistic wound assessment.
    \item \textbf{Advanced Multi-Modal Learning:} Incorporating electronic health record data (comorbidities, lab values) will enable moving from classification to \textit{personalized prognostic forecasting} (e.g., predicting healing time).
    \item \textbf{Embedded Edge Deployment:} Optimizing the ensemble via model quantization and pruning for efficient execution on mobile devices and embedded IoMT hardware is essential for point-of-care utility.
    \item \textbf{Prospective Clinical Validation:} A multi-center randomized controlled trial is being designed to measure the system's impact on key outcomes: healing times, amputation rates, and healthcare costs.
\end{itemize}

\section{Conclusion}
\label{sec:conclusion}
We presented WoundNet-Ensemble, a novel IoMT system that synergistically combines ResNet-50, DINOv2, and Swin Transformer in a weighted ensemble to achieve state-of-the-art accuracy (99.90\%) for automated wound classification across six clinically distinct etiologies. The integration of self-supervised learning addresses medical data limitations, while the ensemble design ensures robustness. The integrated wound healing tracker provides clinicians with quantitative, longitudinal metrics for evidence-based decision-making.

This work demonstrates the significant potential of advanced deep learning ensembles to deliver objective, accurate, and scalable wound assessment tools. By bridging high-performance AI with clinical workflow needs, WoundNet-Ensemble represents a meaningful step toward transforming the standard of care for millions of patients with chronic wounds globally. To promote transparency and reproducibility, the codebase, trained models, and evaluation scripts will be released publicly upon acceptance.

\section*{Acknowledgment}
The authors thank the contributors to the Medetec Wound Database and the AZH Wound Care Dataset. This research was conducted at The Catholic University of America, Department of Electrical Engineering and Computer Science. This research did not receive specific grant funding.



\end{document}